\documentclass[11pt]{article}

\usepackage[final]{acl}

\usepackage{times}
\usepackage{latexsym}

\usepackage[T1]{fontenc}

\usepackage[utf8]{inputenc}

\usepackage{microtype}

\usepackage{inconsolata}

\usepackage{graphicx}

\usepackage{amsmath}
\usepackage{amssymb}
\usepackage{algorithm}
\usepackage{algpseudocode}
\usepackage{booktabs}
\usepackage{tcolorbox}
\usepackage{float}
\usepackage{multirow}
\usepackage{tabularx}
\usepackage{adjustbox}
\usepackage{pifont}
\usepackage{placeins}

\newcommand{\best}[1]{\textbf{#1}}
\newcommand{\second}[1]{\underline{#1}}

\title{CoRA: Confidence--Rationale Alignment for Reliable Chain-of-Thought Reasoning}

\author{
 \textbf{Juming Xiong\textsuperscript{1}},
 \textbf{Weixin Liu\textsuperscript{1}},
 \textbf{Kevin Guo\textsuperscript{1}},
 \textbf{Congning Ni\textsuperscript{2}},
 \textbf{Junchao Zhu\textsuperscript{1}},
 \textbf{Chongyu Qu\textsuperscript{1}},
\\
 \textbf{Chao Yan\textsuperscript{2}},
 \textbf{Katherine Brown\textsuperscript{2}},
 \textbf{Avinash Baidya\textsuperscript{3}},
 \textbf{Xiang Gao\textsuperscript{3}},
 \textbf{Bradley Malin\textsuperscript{1,2}},
 \textbf{Zhijun Yin\textsuperscript{1,2}},
\\
\\
 \textsuperscript{1}Vanderbilt University,
 \textsuperscript{2}Vanderbilt University Medical Center,
 \textsuperscript{3}Intuit AI Research
}

\begin{document}
\maketitle
\begin{abstract}
Chain-of-thought (CoT) reasoning can improve LLM performance, but high answer confidence may be misleading when the accompanying CoT rationale is plausible yet incomplete or poorly supported. We study confidence--rationale alignment: whether a model's confidence in its committed answer is justified by its generated rationale. We introduce a GRPO-based reinforcement learning framework that jointly rewards answer correctness, committed-answer probability, and rubric-based rationale support, where the rubric assesses grounding, coherence, task match, and connection to the selected answer without revealing the gold answer to the judge. Across MedQA, MathQA, and OpenBookQA using three open-weight LLMs, our method reduces the confidence--rationale alignment error by up to 26.51\% compared with untuned checkpoints, SFT, and correctness-only GRPO, while maintaining competitive accuracy and often improving calibration. These results show that reliable CoT reasoning requires not only confident answers, but rationales that substantively support them.
% Chain-of-thought (CoT) reasoning has been shown to improve large language model (LLM) performance. However, an LLM can produce a plausible but incomplete reasoning trace while assigning high probability to the committed answer, creating a misleading reliability signal. We investigate \emph{confidence--rationale alignment}, the extent to which a model's confidence in its committed answer is justified by the generated rationale, i.e., the CoT explanation accompanying the final answer. We introduce a reinforcement learning framework that combines binary answer correctness, rubric-based rationale-support quality, and committed-answer probability in a Group Relative Policy Optimization (GRPO) reward. The rubric evaluates whether a rationale is grounded, coherent, task-matched, and properly connected to the selected answer without exposing the gold answer to the judge. We evaluate our method on three benchmark datasets, MedQA, MathQA, and OpenBookQA, across three open-weight LLMs: Ministral 3 8B Reasoning, Qwen2.5 7B, and Gemma 2 9B Instruct. Compared with the corresponding untuned checkpoints and two training baselines, supervised fine-tuning (SFT) and correctness-only GRPO, our method reduces the confidence--rationale alignment gap by up to 11.1 percentage points across evaluation judges while maintaining competitive accuracy and often improving committed-answer probability calibration metrics. These findings suggest that reliable reasoning should be evaluated not only by confidence but also by the degree to which the rationale supports the committed answer.
\end{abstract}

\section{Introduction}

\begin{figure}[t]
\centering
\includegraphics[width=1.0\linewidth]{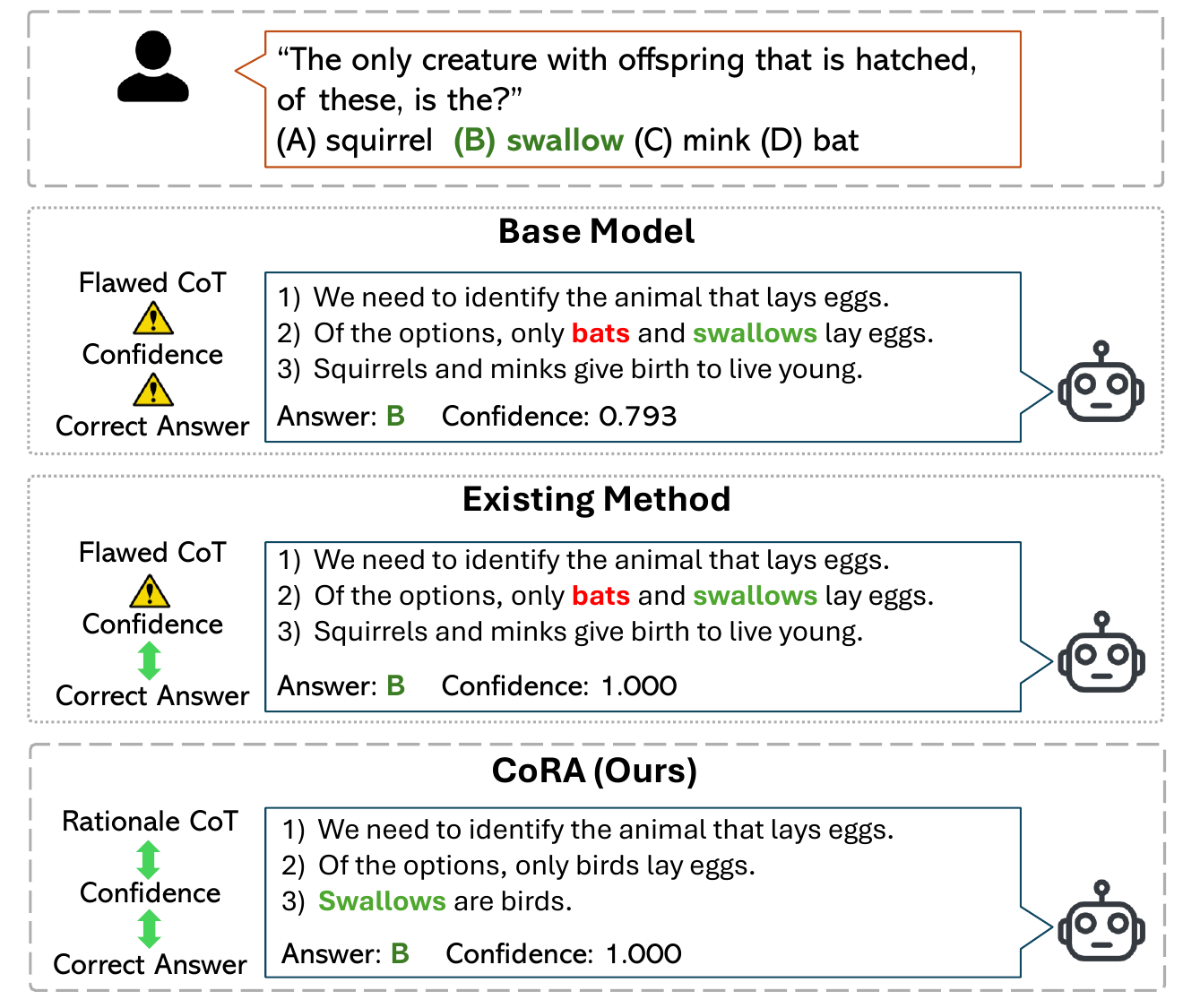}
\caption{
An example of confidence--rationale misalignment. 
Both the base model and a correctness-oriented method choose the correct answer with near-perfect confidence but include an unsupported claim that bats lay eggs.
Our method produces a rationale that better supports the committed answer by connecting the selected option, swallow, to hatched offspring.
}
\label{fig:example}
\end{figure}

Chain-of-thought (CoT) reasoning has been shown to improve large language model (LLM) performance on arithmetic, commonsense, symbolic, and other reasoning tasks \citep{wei2022chain,kojima2022large}. In this paper, we use \emph{rationale} to refer to the generated CoT explanation that accompanies the final answer. In many high-stakes user-facing settings, the reliability of a response is judged not only by whether the final answer is correct, but also by whether the model appears confident and whether its rationale justifies the answer. This motivates \emph{confidence--rationale alignment}: the extent to which a model's confidence in its committed answer is justified by the rationale it generates.

This problem is important because a model may produce a fluent rationale, choose an incorrect answer, and still assign high confidence to that answer. Such failures can be subtle: when an answer is accompanied by both a persuasive rationale and a strong confidence signal, users may find the error difficult to recognize or correct. Prior work in human-AI interaction shows that rationales and explanations can strongly shape user reliance on model outputs, sometimes increasing reliance even when the response is incorrect \citep{vasconcelos2023explanations,steyvers2025what,kim2025fostering}. A trustworthy reasoning model should therefore be confident only when its rationale adequately supports the answer it commits to.

Existing work addresses related parts of this problem, but not their interaction. CoT prompting, self-consistency, and tree-structured search improve reasoning accuracy by eliciting or searching over intermediate reasoning traces \citep{wei2022chain,wang2023selfconsistency,yao2023tree}. However, generated rationales may not faithfully reflect the factors that determine the model's answer \citep{turpin2023unfaithful,lanham2023faithfulness,paul2024making,tutek2025measuring}. Separately, calibration methods aim to make answer confidence reflect empirical correctness \citep{guo2017calibration,zhao2021calibrate,kadavath2022language,tian2023just,xie2024calibrating,thermometer2024}. Yet they typically evaluate output probabilities in aggregate and do not assess whether an individual rationale supports the answer at the matched confidence level. This distinction matters in human-facing systems, where confidence and explanation signals can affect user reliance and may not be interpreted uniformly across users \citep{lin2022teaching,tian2023just,steyvers2025what,kim2025fostering}.

To address this gap, we introduce a confidence--rationale alignment framework, CoRA, for multiple-choice reasoning. CoRA has two components. First, we use a structured LLM-as-judge rubric to assess whether a rationale is grounded, coherent, task-matched, and properly connected to the model's committed answer without exposing the gold answer \citep{zheng2023judging,liu2023geval,kim2024prometheus}. Second, we optimize the LLM with a Group Relative Policy Optimization (GRPO)-based reward \citep{shao2024deepseekmath,deepseekai2025deepseekr1arxiv} that combines answer correctness, rationale-support quality, and committed-answer confidence. Unlike correctness-only GRPO, this objective encourages the model to ground its confidence in the generated rationale rather than treating confidence as an isolated scalar.

We evaluate CoRA on three benchmark datasets, MedQA, MathQA, and OpenBookQA \citep{jin2021medqa,amini2019mathqa,mihaylov2018openbookqa}, across three open-weight models. We measure answer accuracy, committed-answer probability calibration using expected calibration error (ECE) and Brier score \citep{brier1950verification}, and confidence--rationale mismatch, which captures cases where confidence exceeds rationale-support quality. Empirically, CoRA most consistently reduces unsupported overconfidence, with the clearest gains on MathQA, while maintaining competitive accuracy.

Our contributions are fourfold. 
(1) We formulate \emph{confidence--rationale alignment} as a reliability problem for reasoning LLMs, requiring committed-answer confidence to be justified by the generated rationale. 
(2) We design a structured LLM-as-judge rubric that evaluates rationale support for the model's selected answer without exposing the gold answer. 
(3) We propose a GRPO-based training framework that combines answer correctness, rationale-support quality, and committed-answer confidence to reduce unsupported overconfidence. 
(4) We evaluate CoRA on three benchmark datasets and three open-weight models, showing reduced confidence--rationale error in most settings while maintaining competitive accuracy; we further introduce a downstream correctness-prediction task showing that CoRA can make generated reasoning traces more diagnostically informative.

\section{Related Work}

\subsection{CoT Reasoning and Rationale Faithfulness}

CoT prompting improves LLM reasoning by eliciting intermediate reasoning steps before a final answer \citep{wei2022chain}. Follow-up methods such as zero-shot CoT, self-consistency, Tree of Thoughts, and STaR further show that generated reasoning traces can improve task accuracy through prompting, sampling, search, or bootstrapping from model-generated rationales \citep{kojima2022large,wang2023selfconsistency,yao2023tree,zelikman2022star}. These methods demonstrate the practical value of intermediate reasoning traces, but they primarily optimize or evaluate final-answer performance.

Improved reasoning performance, however, does not guarantee that a generated rationale faithfully explains the model's prediction. Prior work shows that CoT rationales can omit factors that influence model outputs, rationalize biased or incorrect answers, or fail to causally determine the final answer \citep{turpin2023unfaithful,lanham2023faithfulness}. More recent work uses causal mediation, process verification, or unlearning-based interventions to test whether reasoning steps influence or justify final predictions \citep{lightman2023lets,paul2024making,tutek2025measuring}. Our work builds on this reliability concern, but focuses on a different question: whether the model's confidence in its committed answer is justified by the rationale it presents.

\subsection{Confidence Calibration and Confidence--Quality Alignment}

Confidence calibration aims to ensure that predictive confidence reflects empirical correctness. Classical calibration work shows that neural networks can be poorly calibrated and that post-hoc methods such as temperature scaling can improve probability estimates \citep{guo2017calibration}. In LLMs, prior work studies prompt-induced calibration errors, whether models can assess the correctness of their own answers, and how models express uncertainty in probabilities or words \citep{zhao2021calibrate,kadavath2022language,lin2022teaching,tian2023just}. Other work proposes post-hoc or auxiliary calibration methods and surveys broader confidence estimation techniques for LLMs \citep{xie2024calibrating,thermometer2024,geng2024survey}.

Recent work has also begun to connect confidence with response quality. CONQORD aligns verbalized confidence with response quality using reinforcement learning \citep{conqord2024}, while CoT-UQ and CER study whether CoT or confidence signals can improve uncertainty quantification and reasoning behavior \citep{zhang2025cot,razghandi2025cer}. These studies are closely related to our motivation, but our setting differs in two ways: 1) we use committed-answer probability rather than verbalized confidence; 2) we explicitly evaluate whether confidence is supported by the rationale, yielding an instance-level alignment signal beyond aggregate calibration.

\subsection{Rationale Evaluation, LLM-as-Judge, and RL Training}

Our rationale-support rubric is grounded in prior work on rationales, faithfulness, and rubric-based evaluation. These studies investigate how models can provide textual evidence for their decisions \citep{lei2016rationalizing} and provide a benchmark for evaluating rationalized NLP models \citep{deyoung2020eraser}. \citet{jacovi2020towards} further emphasizes that explanations should be evaluated by their relationship to model predictions, not only by their surface plausibility. These ideas motivate our focus on answer-support quality: whether a rationale uses relevant evidence, follows coherent inference steps, and bridges to the model's selected answer.

LLM-as-judge evaluation provides a scalable way to assess open-ended outputs, but prior work also shows that judge behavior can be sensitive to rubric design and may exhibit biases \citep{zheng2023judging}. G-Eval and Prometheus demonstrate that structured evaluation prompts and fine-grained rubrics can improve the consistency and usefulness of model-based evaluation \citep{liu2023geval,kim2024prometheus}. Following this line of work, our judge is constrained by a structured rubric and is not given the gold answer, so that it evaluates whether the rationale supports the model's own committed answer rather than whether the answer is correct.

Our optimization method is related to reinforcement learning for LLMs. PPO-style optimization has been widely used in LLM alignment \citep{schulman2017proximal,ouyang2022training}, and recent reasoning-oriented RL methods such as DeepSeekMath and DeepSeek-R1 show that reinforcement learning can improve mathematical and general reasoning behavior \citep{shao2024deepseekmath,deepseekai2025deepseekr1arxiv}. In contrast to correctness-only reinforcement learning, our reward incorporates answer correctness, rationale-support quality, and committed-answer confidence. This connects reasoning supervision with calibration by training models to reduce unsupported overconfidence rather than only maximizing final-answer correctness.

\section{Method}

\begin{figure*}[t]
\centering
\includegraphics[width=0.8\textwidth]{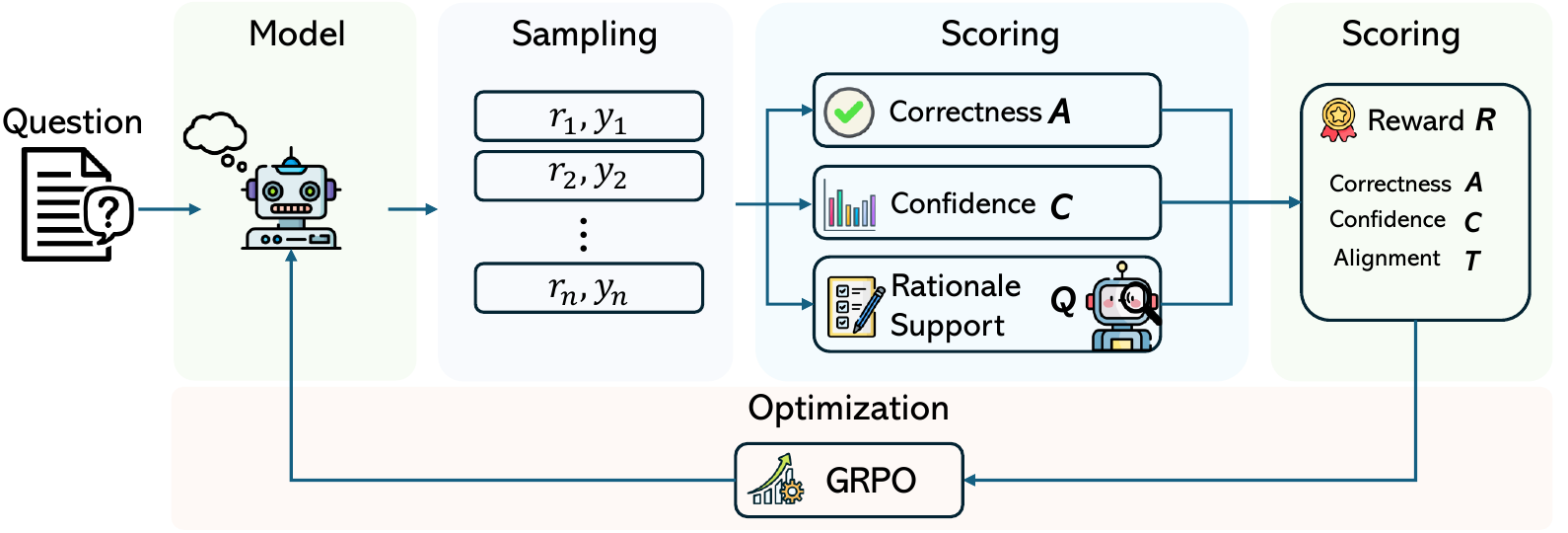}
\caption{
\textbf{An overview of the CoRA framework.} Given multiple-choice questions, the policy model samples a group of responses, each consisting of a rationale and a committed answer. For each response, an LLM judge estimates rationale-support quality \(Q\), while the policy model provides committed-answer confidence \(C\). We first combine \(Q\) and \(C\) into an alignment score \(T\), and then use \(T\), correctness \(A\), and confidence \(C\) to compute the final reward \(R\) for GRPO optimization. The reward favors well-supported confident answers and penalizes unsupported overconfidence.
}
\label{fig:framework}
\end{figure*}

We develop a reinforcement learning framework for confidence--rationale alignment in multiple-choice reasoning, as shown in Figure~\ref{fig:framework}. %Given a question \(x\), a set of answer options \(\mathcal{Y}\), a generated rationale \(r\), and a committed answer \(y\in\mathcal{Y}\), our objective is to train a model whose confidence in \(y\) is grounded in the support provided by \(r\). The framework has three components: a modular rubric for estimating rationale-support quality, a confidence--rationale alignment score, and a GRPO training objective.

\subsection{Rubric for Assessing Rationale Quality}

We define rationale quality as the extent to which a generated rationale provides coherent, grounded, and task-appropriate support for the model's committed answer. We estimate this quality using an LLM-as-judge that takes as input the dataset name, question, answer options, optional auxiliary background knowledge, generated rationale, and committed answer. The auxiliary background knowledge does not include the gold option label, official solution, official explanation, or gold rationale. The judge evaluates whether the rationale supports the committed answer, not whether the committed answer is correct. We treat this judge score as a scalable proxy for rationale-support quality rather than as a human gold standard. To reduce label leakage, the judge is constrained by a structured rubric and is not given the gold answer.

The rubric is designed based on criteria that recur in prior work, which emphasizes 1) whether explanations identify relevant evidence and support model predictions \citep{lei2016rationalizing,deyoung2020eraser,jacovi2020towards}; 2) CoT faithfulness work highlights the need to assess whether reasoning steps coherently justify final predictions \citep{turpin2023unfaithful,lanham2023faithfulness,paul2024making}; and 3) LLM-as-judge work motivates explicit, fine-grained rubrics to reduce ambiguity in open-ended evaluation \citep{zheng2023judging,liu2023geval,kim2024prometheus}. Based on these principles, our general axes cover format validity, task understanding, evidence grounding, inference coherence, answer bridging, and structure. The task-specific axes add dataset-relevant requirements, such as computation and quantity setup for MathQA, clinical application for MedQA, and science-fact application for OpenBookQA.

The rubric provides an operational measure of answer-support quality. It evaluates whether a generated reasoning trace supports the model's committed answer through grounded, coherent, and task-matched reasoning. To reduce judge-specific bias, we evaluate alignment with three independent judge models and report whether the observed trends remain consistent across judges.

\begin{table}[t]
\centering
\footnotesize
\setlength{\tabcolsep}{4pt}
\renewcommand{\arraystretch}{1.05}
\begin{tabular}{lll}
\toprule
\textbf{Group} & \textbf{Axis} & \textbf{Description} \\
\midrule
\multirow{6}{*}{General}
& \(G_1\) & Format validity \\
& \(G_2\) & Task understanding \\
& \(G_3\) & Evidence grounding \\
& \(G_4\) & Inference coherence \\
& \(G_5\) & Answer bridge \\
& \(G_6\) & Structure \\
\midrule
\multirow{2}{*}{OpenBookQA}
& \(O_1\) & Science fact correctness \\
& \(O_2\) & Scenario application \\
\midrule
\multirow{2}{*}{MedQA}
& \(\mathrm{Med}_1\) & Medical fact correctness \\
& \(\mathrm{Med}_2\) & Clinical application \\
\midrule
\multirow{2}{*}{MathQA}
& \(\mathrm{Math}_1\) & Computation correctness \\
& \(\mathrm{Math}_2\) & Quantity setup \\
\bottomrule
\end{tabular}
\caption{A modular rationale-support rubric. General axes are shared across datasets; each dataset adds two task-specific axes.}
\label{tab:rubric}
\end{table}

Each axis is assigned a categorical label and mapped to a scalar value in \([0,1]\). Let \(Q_{\mathrm{gen}}\) denote the aggregate score over the six general axes, and \(Q_{\mathrm{task}}\) denote the aggregate score over the two task-specific axes for the corresponding dataset. The final rationale-support score is:
\begin{equation}
Q = w_{\mathrm{gen}} Q_{\mathrm{gen}}
  + w_{\mathrm{task}} Q_{\mathrm{task}} - P,
\label{eq:q-score}
\end{equation}
where \(P\) denotes a penalty for severe structural failure or unsupported reasoning. The resulting score is clipped to \([0,1]\). The format axis \(G_1\) acts as a gate: if the response lacks a valid parseable answer, \(Q\) is set to zero.

\subsection{Confidence--Rationale Alignment Reward}

We design a reward function that jointly considers answer correctness, rationale-support quality, and committed-answer confidence.

\paragraph{Committed-answer confidence.}
Let \(\mathcal{Y}\) denote the set of answer options and let \(y\in\mathcal{Y}\) be the parsed answer generated by the model. We compute the committed-answer confidence \(C\) by normalizing model scores over candidate answer labels:
\begin{equation}
C =
\frac{\exp(s_y)}
{\sum_{\ell\in\mathcal{Y}}\exp(s_\ell)} ,
\label{eq:confidence}
\end{equation}
where \(s_\ell\) is the conditional log-probability score of candidate label \(\ell\) under the exact generated prefix up to, but excluding, the final answer label. Thus, \(C\in[0,1]\) measures the normalized probability assigned to the model's committed answer. We use option-label probabilities rather than verbalized confidence. Details are described in Appendix~\ref{app:confidence-extraction}.

\paragraph{Alignment score.}
A desirable response should maintain consistency between confidence and rationale support: higher confidence should be accompanied by stronger rationale support, while weakly supported rationales should receive lower confidence. We operationalize this criterion with a confidence--rationale alignment score \(T\) as follows:
\begin{equation}
\begin{aligned}
T = \operatorname{clip}\big(&
\alpha Q + (1-\alpha)C \\
&- \beta_{\mathrm{over}}\max(C-Q,0) \\
&- \beta_{\mathrm{under}}\max(Q-C,0),
0,1 \big).
\end{aligned}
\label{eq:t-score}
\end{equation}
The first two terms combine rationale support and committed-answer confidence. The penalty terms capture the extent of the confidence--rationale mismatch: \(\max(C-Q,0)\) penalizes confidence that exceeds rationale support, while \(\max(Q-C,0)\) penalizes confidence that falls below rationale support. We use asymmetric penalties to prioritize reducing unsupported high-confidence answers.

\paragraph{Overall reward.}
The final reward combines correctness and confidence--rationale alignment:
\begin{equation}
\begin{aligned}
R =\;& A(1+\lambda T)
- \gamma(1-A)C \\
&- \beta \max(C-Q,0).
\end{aligned}
\label{eq:reward}
\end{equation}
Here, \(A\in\{0,1\}\) is a binary answer-correctness signal, while \(Q\), \(C\), and \(T\) are continuous scores in \([0,1]\). The first term rewards correct answers and those with stronger confidence--rationale alignment. The second term penalizes confident wrong answers, while the third term penalizes confidence that exceeds rationale support, regardless of answer correctness. This reward encourages the model to produce accurate answers, generate rationales that support those answers, and calibrate its confidence according to the quality of its reasoning.

\subsection{Optimization}

We optimize the policy model using GRPO. For each input question, the policy samples a group of candidate completions. Each completion is parsed into a rationale and a committed answer, scored for answer correctness, evaluated by the LLM judge for rationale-support quality, and assigned a committed-answer confidence score. The resulting reward in Equation~\ref{eq:reward} is applied to compute group-relative advantages within the sampled response group. The policy is then updated to increase the likelihood of responses with higher relative rewards while regularizing against large deviations from the reference policy. This allows the model to learn from multiple signals simultaneously. % Correct answers with stronger rationale support and better confidence--rationale alignment receives a larger reward. Incorrect answers are penalized more strongly when the model assigns high confidence, and responses whose confidence exceeds the support provided by the rationale are penalized regardless of correctness.

\subsection{Correctness Prediction from CoT Reasoning Linguistic Features}

As an auxiliary diagnostic of user-facing reasoning transparency, we test whether surface linguistic features of generated reasoning traces contain information about answer correctness. Given a generated reasoning trace \(r_i\), we extract a feature vector \(\phi(r_i)\) and train a logistic regression classifier to predict the binary correctness label. We use stratified cross-validation and report out-of-fold predicted probabilities. Feature standardization is performed within each training fold to avoid information leakage.
The full feature set is listed in Appendix~\ref{app:linguistic-feature} Table~\ref{tab:linguistic-features}.

\section{Experiments}

\subsection{Experimental Settings}

\paragraph{Datasets.}
We evaluate CoRA on three multiple-choice reasoning benchmarks spanning mathematical, medical, and scientific reasoning.

\begin{itemize}
    \item \textbf{MathQA} \cite{amini2019mathqa}: A mathematical reasoning benchmark consisting of multiple-choice word problems that require quantity identification, symbolic reasoning, and arithmetic computation.

    \item \textbf{MedQA} \cite{jin2021medqa}: A medical question-answering benchmark derived from medical licensing exams. It evaluates clinical and biomedical reasoning over patient-centered and factual medical scenarios.

    \item \textbf{OpenBookQA} \cite{mihaylov2018openbookqa}: An elementary science question-answering benchmark that requires using scientific facts and commonsense knowledge for selecting the correct answer.
\end{itemize}

\paragraph{Models.}
We use three open-weight models as policy models: Ministral 3 8B Reasoning \cite{mistral2025ministral3reasoning}, Qwen2.5 7B \cite{qwen2025qwen25technicalreport}, and Gemma 2 9B Instruct \cite{gemmateam2024gemma2}. These models cover different model families and instruction-tuning settings. We use the same prompting format and evaluation protocol across all policy models and datasets.

For rubric-based training, we use GPT-OSS 20B as the judge model. The judge is provided with task-relevant auxiliary background knowledge, but not the gold option label, official solution, official explanation, or gold rationale. This setup approximates a stronger evaluator with access to external evidence, reducing the chance that the judge rewards or penalizes a reasoning trace due to missing background knowledge.

For evaluation, we use three judge models, GPT-OSS 20B~\citep{agarwal2025gpt}, Llama 3.1 70B Instruct~\citep{dubey2024llama}, and Qwen 3 32B~\citep{yang2025qwen3technicalreport} to measure rationale-support quality and confidence--rationale alignment. The evaluation judges are also provided with the same auxiliary background knowledge used by the training judge, under the same no-gold-information constraint. This design allows us to assess whether the observed alignment trends remain stable across different judge models rather than depending on a single evaluator.

\paragraph{Baselines.}
We compare CoRA with three training baselines: 1) \textbf{Base}: The original model without task-specific fine-tuning; 2) \textbf{SFT}: Supervised fine-tuning on task examples containing rationales and final answers; 3) \textbf{GRPO}: GRPO training with a rule-based correctness reward, where the reward is determined by whether the parsed answer matches the gold label. It uses the same optimization setup as CoRA, but replaces the confidence--rationale reward with a correctness-only reward.

\subsection{Evaluation Metrics}

To evaluate whether CoRA improves both task performance and confidence--rationale alignment, we employ accuracy and calibration metrics, together with rationale-centered alignment diagnostics.

\begin{itemize}
    \item \textbf{Accuracy}: It measures whether the parsed answer matches the gold label. We use the final parsed option as the model prediction.

    \item \textbf{Expected Calibration Error (ECE)} \cite{guo2017calibration}: ECE measures the discrepancy between committed-answer confidence and empirical correctness. We compute ECE using 10 equal-width confidence bins.

    \item \textbf{Brier Score} \cite{brier1950verification}: It is a proper scoring rule for probabilistic calibration. For each example, it computes the squared difference between committed-answer confidence and binary correctness. % Lower Brier score indicates better calibration.

    \item \textbf{Rationale-Support Quality}: We evaluate it using a rubric-based judge, which assigns each generated reasoning trace a score \(Q\in[0,1]\). It measures whether the reasoning trace is grounded, coherent, task-matched, and supportive of the committed answer. % Higher values indicate stronger rationale support.
    
    \item \textbf{Confidence--Rationale Alignment Error}:
    We evaluate confidence--rationale alignment using the average absolute gap between committed-answer confidence and rationale-support quality:
    \begin{equation}
    E_{\mathrm{align}}
    =
    \frac{1}{N}
    \sum_{i=1}^{N}
    |C_i-Q_i| .
    \end{equation}
    Lower \(E_{\mathrm{align}}\) indicates stronger alignment between confidence and rationale support. 
    % \item \textbf{Rationale-Support Quality}: We introduce a rubric-based judge to assign each generated reasoning trace a rationale-support score \(Q\in[0,1]\). The score evaluates whether the reasoning trace is grounded, coherent, task-matched, and supportive of the committed answer. Higher values indicate stronger rationale support.
    % \item \textbf{Confidence--Rationale Alignment Gap}: We evaluate alignment between committed-answer confidence and rationale-support quality using:
    % \begin{equation}
    % \Delta_{\mathrm{align}}
    % =
    % \frac{1}{N}
    % \sum_{i=1}^{N}
    % |C_i-Q_i| ,
    % \end{equation}
    % where \(C_i\) is the committed-answer confidence and \(Q_i\) is the rationale-support score for instance \(i\). Lower values indicate better confidence--rationale alignment. %We compute this metric with GPT-OSS 20B as the main judge and Llama 3 70B as an additional judge for robustness analysis.
\end{itemize}

\subsection{Implementation Details}

For each dataset, we use 200 samples for training and 500 held-out samples for evaluation. For SFT, the 200 training examples are selected from instances that the corresponding base model answers correctly. For GRPO and CoRA, we randomly sample 200 examples from the training set. The same evaluation subset is used across all methods for a controlled comparison. During GRPO-based training, we sample 8 completions per question with a sampling temperature of 0.8.

\section{Results}

\begin{table*}[t]
\centering
\caption{\textbf{Results on accuracy and calibration scores on datasets and models.} Accuracy, ECE, and Brier score are reported as percentages. Higher accuracy is better, while lower ECE and Brier score indicate better calibration. Best values are bolded and second-best values are underlined within each model--dataset block.}
\scriptsize
\setlength{\tabcolsep}{2pt}
\renewcommand{\arraystretch}{1.0}
\resizebox{\textwidth}{!}{
\begin{tabular}{llccc@{\hspace{2em}}ccc@{\hspace{2em}}ccc}
\toprule
& & \multicolumn{3}{c}{\textbf{Ministral 3 8B}} 
& \multicolumn{3}{c}{\textbf{Qwen2.5 7B}} 
& \multicolumn{3}{c}{\textbf{Gemma 2 9B Instruct}} \\
\cmidrule(lr){3-5}
\cmidrule(lr){6-8}
\cmidrule(lr){9-11}
\textbf{Dataset} & \textbf{Method}
& \textbf{Acc} $\uparrow$ & \textbf{ECE} $\downarrow$ & \textbf{Brier} $\downarrow$
& \textbf{Acc} $\uparrow$ & \textbf{ECE} $\downarrow$ & \textbf{Brier} $\downarrow$
& \textbf{Acc} $\uparrow$ & \textbf{ECE} $\downarrow$ & \textbf{Brier} $\downarrow$ \\
\midrule
\multirow{4}{*}{MathQA}
& Base & 79.40 & 8.35 & 26.43 & 72.00 & \second{11.55} & 37.69 & 58.80 & 32.21 & 68.08 \\
& SFT  & \best{80.20} & 12.40 & 29.51 & 72.80 & 17.55 & 41.69 & 61.20 & 34.43 & 70.67 \\
& GRPO & \second{79.80} & \second{7.88} & \second{25.86} & \best{77.20} & 13.94 & \second{35.50} & \second{61.80} & \second{29.90} & \second{64.70} \\
& CoRA (Ours) & \best{80.20} & \best{6.58} & \best{22.66} & \second{76.40} & \best{11.09} & \best{32.83} & \best{63.40} & \best{25.60} & \best{58.27} \\
\midrule
\multirow{4}{*}{MedQA}
& Base & 65.00 & \second{31.56} & 65.73 & 46.60 & 42.94 & 93.13 & 51.80 & 46.86 & 93.86 \\
& SFT  & \second{65.20} & 33.06 & 67.83 & 50.40 & 44.45 & 91.39 & 52.00 & 47.58 & 94.83 \\
& GRPO & \best{66.80} & \best{30.37} & \best{62.66} & \best{54.40} & \best{35.17} & \best{76.74} & \second{54.60} & \second{43.28} & \second{87.56} \\
& CoRA (Ours) & 64.60 & 31.91 & \second{65.41} & \second{52.00} & \second{39.72} & \second{84.79} & \best{55.60} & \best{42.22} & \best{86.48} \\
\midrule
\multirow{4}{*}{OpenBookQA}
& Base & 83.40 & 13.94 & 30.50 & 80.80 & \best{10.65} & 30.45 & 78.80 & 19.16 & 38.77 \\
& SFT  & \second{87.20} & \best{11.54} & \second{24.23} & \second{82.80} & 12.71 & 29.74 & 80.60 & 19.05 & 37.86 \\
& GRPO & 86.40 & \best{11.54} & 24.78 & 82.60 & \second{10.73} & \second{27.80} & \best{84.20} & \second{14.52} & \second{30.23} \\
& CoRA (Ours) & \best{87.40} & 11.72 & \best{23.96} & \best{86.00} & 11.89 & \best{25.57} & \second{84.00} & \best{14.14} & \best{30.21} \\
\bottomrule
\end{tabular}
}

\label{tab:main-results}
\end{table*}

\subsection{Accuracy and Calibration Scores.}

We conduct experiments to evaluate the effectiveness of CoRA on MathQA, MedQA, and OpenBookQA across three foundation models. Table~\ref{tab:main-results} reports answer accuracy, Expected Calibration Error (ECE), and Brier score. % Ideally, a reliable model should achieve high accuracy while maintaining low ECE and Brier score, indicating that its committed-answer confidence is well aligned with empirical correctness.

We observe that CoRA generally improves calibration while preserving competitive task performance. In particular, CoRA achieves the best Brier score in most model--dataset settings, including all three datasets for Gemma 2 9B Instruct, MathQA and OpenBookQA for Ministral 3 8B, and MathQA and OpenBookQA for Qwen2.5 7B. This suggests that incorporating confidence--rationale alignment into reinforcement learning improves the probabilistic reliability of the committed answers.

CoRA also maintains relatively strong answer accuracy. On MathQA, CoRA obtains the best or tied-best accuracy for Ministral 3 8B and Gemma 2 9B Instruct, while remaining competitive with GRPO on Qwen2.5 7B. On OpenBookQA, CoRA achieves the best accuracy for Ministral 3 8B and Qwen2.5 7B, and the second-best accuracy for Gemma 2 9B Instruct. These results indicate that optimizing for confidence--rationale alignment does not require sacrificing task accuracy.

% Compared with the GRPO baseline, which optimizes only answer correctness, CoRA often provides better calibration. For example, on MathQA, CoRA reduces both ECE and Brier score across all three models. On Gemma 2 9B Instruct, CoRA consistently improves ECE and Brier score over GRPO on all datasets. This demonstrates that answer-only reinforcement learning is not sufficient to ensure reliable confidence behavior, whereas CoRA directly encourages confidence to be grounded in the generated rationale.

However, CoRA is not uniformly best in every setting. On MedQA, GRPO achieves the strongest calibration for Ministral 3 8B and Qwen2.5 7B, while CoRA performs best for Gemma 2 9B Instruct. This suggests that the benefit of confidence--rationale alignment depends on the interaction between model family and domain. Nevertheless, CoRA remains consistently competitive and often improves calibration-sensitive metrics.

\begin{table}[t]
\centering
\caption{\textbf{Rationale-support quality and confidence--rationale alignment results evaluated by Llama 3.1 70B Instruct.} \(Q\) measures the quality of the generated reasoning trace, while \(E_{\mathrm{align}}\) denotes the average confidence--rationale alignment error. \(\uparrow\) indicates higher is better, and \(\downarrow\) indicates lower is better. All values are reported as percentages.}
\scriptsize
\setlength{\tabcolsep}{2pt}
\renewcommand{\arraystretch}{1.03}
\resizebox{\columnwidth}{!}{
\begin{tabular}{llcccccc}
\toprule
\multicolumn{2}{c}{} 
& \multicolumn{2}{c}{\textbf{Ministral 3 8B}}
& \multicolumn{2}{c}{\textbf{Qwen2.5 7B}}
& \multicolumn{2}{c}{\textbf{Gemma 2 9B}} \\
\cmidrule(lr){3-4} \cmidrule(lr){5-6} \cmidrule(lr){7-8}
\textbf{Dataset} & \textbf{Method}
& \(Q\) $\uparrow$ & \(E_{\mathrm{align}}\) $\downarrow$
& \(Q\) $\uparrow$ & \(E_{\mathrm{align}}\) $\downarrow$
& \(Q\) $\uparrow$ & \(E_{\mathrm{align}}\) $\downarrow$ \\
\midrule
\multirow{4}{*}{MathQA}
& Base & \second{97.56} & 4.56 & 86.55 & 14.35 & 82.75 & 16.70 \\
& SFT  & 96.75 & \second{3.92} & 87.52 & 12.88 & \second{84.01} & 15.89 \\
& GRPO & 97.12 & 4.55 & \second{87.82} & \second{10.28} & 83.86 & \second{15.73} \\
& CoRA (Ours) & \best{98.18} & \best{1.65} & \best{91.57} & \best{9.53} & \best{89.96} & \best{11.28} \\
\midrule
\multirow{4}{*}{MedQA}
& Base & 94.90 & 6.09 & 73.42 & 26.20 & 81.47 & 18.31 \\
& SFT  & 94.88 & \second{5.35} & \second{77.67} & \second{21.76} & 82.59 & 17.14 \\
& GRPO & \best{95.65} & 5.79 & 74.77 & 22.31 & \second{82.63} & \second{16.91} \\
& CoRA (Ours) & \second{94.99} & \best{5.08} & \best{79.87} & \best{20.29} & \best{94.47} & \best{6.02} \\
\midrule
\multirow{4}{*}{OpenBookQA}
& Base & 95.34 & 5.28 & \second{91.94} & 10.44 & 90.38 & 9.02 \\
& SFT  & \best{97.58} & \best{2.58} & 91.92 & \second{8.48} & 90.54 & 9.06 \\
& GRPO & 95.91 & 4.59 & 90.90 & 9.46 & \second{92.14} & \second{7.59} \\
& CoRA (Ours) & \second{96.09} & \second{3.66} & \best{94.65} & \best{5.65} & \best{92.60} & \best{7.29} \\
\bottomrule
\end{tabular}
}
\label{tab:q-align-results-llama}
\end{table}

\subsection{Rationale-Support Quality and Alignment}
Table~\ref{tab:q-align-results-llama} reports reasoning quality \(Q\) and alignment error \(E_{\mathrm{align}}\) evaluated by Llama 3.1 70B Instruct. CoRA shows the clearest gains on MathQA, achieving the highest \(Q\) and lowest \(E_{\mathrm{align}}\) across all three policy models. This indicates that CoRA improves both the quality of mathematical reasoning traces and the consistency between rationale support and committed-answer confidence.

On MedQA, CoRA also performs strongly. It obtains the lowest \(E_{\mathrm{align}}\) across all three models and the highest \(Q\) for Qwen2.5 7B and Gemma 2 9B, while GRPO achieves the highest \(Q\) for Ministral 3 8B. On OpenBookQA, CoRA achieves the best results for Qwen2.5 7B and Gemma 2 9B, while SFT performs best for Ministral 3 8B.

Overall, the results show that CoRA generally improves confidence--rationale alignment and reasoning-support quality, with especially consistent gains on MathQA.

To reduce the impact of judge-specific bias in LLM-based evaluation, we further evaluate \(Q\) and \(E_{\mathrm{align}}\) using two additional judge models: GPT-OSS 20B and Qwen 3 32B. These judges follow the same rubric and receive the same auxiliary background knowledge as the main evaluator, under the same no-gold-information constraint. The additional results are reported in Appendix~\ref{app:judge-robustness}, from Table~\ref{tab:q-align-results-gpt} to Table~\ref{tab:q-align-results-qwen3}. Across judges, the overall trends remain broadly consistent, suggesting that the observed improvements are not an artifact of a single evaluator.

\begin{table}[t]
\centering
\caption{\textbf{Prediction Results from CoT Reasoning Linguistic Features.} \(\uparrow\) indicates higher is better. Best values are bolded and second-best values are underlined within each model--dataset block.}
\scriptsize
\setlength{\tabcolsep}{2pt}
\renewcommand{\arraystretch}{1.0}
\resizebox{\columnwidth}{!}{
\begin{tabular}{llccc}
\toprule
\multicolumn{2}{c}{} 
& \textbf{Ministral 3 8B}
& \textbf{Qwen2.5 7B}
& \textbf{Gemma 2 9B} \\
\cmidrule(lr){3-3} \cmidrule(lr){4-4} \cmidrule(lr){5-5}
\textbf{Dataset} & \textbf{Method}
& \textbf{AUROC} $\uparrow$
& \textbf{AUROC} $\uparrow$
& \textbf{AUROC} $\uparrow$ \\
\midrule
\multirow{4}{*}{MathQA}
& Base & \second{0.820} & \second{0.665} & \second{0.694} \\
& SFT  & 0.787 & 0.651 & 0.662 \\
& GRPO & 0.833 & 0.592 & 0.670 \\
& CoRA (Ours) & \textbf{0.846} & \textbf{0.741} & \textbf{0.695} \\
\midrule
\multirow{4}{*}{MedQA}
& Base & \second{0.588} & 0.552 & 0.537 \\
& SFT  & 0.578 & \second{0.555} & \second{0.538} \\
& GRPO & \textbf{0.613} & 0.539 & 0.503 \\
& CoRA (Ours) & {0.572} & \textbf{0.591} & \textbf{0.554} \\
\midrule
\multirow{4}{*}{OpenBookQA}
& Base & \second{0.610} & 0.495 & \textbf{0.560} \\
& SFT  & 0.567 & 0.525 & 0.479 \\
& GRPO & 0.568 & \second{0.542} & 0.489 \\
& CoRA (Ours) & \textbf{0.627} & \textbf{0.581} & \second{0.503} \\
\bottomrule
\end{tabular}
}

\label{tab:linguistic-prediction-auroc}
\end{table}

\subsection{Correctness Prediction from CoT Reasoning
Linguistic Features}
Table~\ref{tab:linguistic-prediction-auroc} evaluates whether surface linguistic features in generated reasoning traces can predict answer correctness. Overall, the results show that linguistic cues provide a useful but limited signal. Many AUROC values on MedQA and OpenBookQA remain close to the weak-discrimination range of 0.50--0.60, indicating that surface linguistic features alone are insufficient for reliable correctness prediction in these settings.

The clearest signal appears on MathQA. Across all models, MathQA obtains substantially higher AUROC than other datasets, and CoRA achieves the best AUROC for every model. For example, on Qwen2.5, CoRA improves linguistic-only AUROC from 0.592 under GRPO to 0.741. This suggests that mathematical reasoning traces contain stronger observable correctness signals, which CoRA further enhances.

The results on MedQA and OpenBookQA are more mixed. CoRA achieves the best AUROC for Qwen2.5 and Gemma 2 on MedQA, and for Ministral 3 and Qwen2.5 on OpenBookQA, but several scores remain near 0.55. This suggests that correctness in medical and science QA may depend more on latent domain knowledge than on shallow linguistic patterns alone. The main takeaway is that CoRA can make generated reasoning traces more diagnostically informative, especially on MathQA.

\section{Discussion and Conclusion}

The results suggest that confidence--rationale alignment is most useful as a reliability-oriented objective rather than as a direct substitute for correctness-oriented training. CoRA most consistently reduces unsupported overconfidence, as measured by the alignment error \(E_{\mathrm{align}}\), while maintaining competitive accuracy. This distinction is important because a model can be accurate or calibrated on average and still produce individual responses whose confidence is not supported by the rationale it presents. Confidence--rationale alignment therefore provides an instance-level reliability signal that complements conventional accuracy and calibration metrics.

The main trade-off is that CoRA improves rationale-centered alignment more consistently than it improves all conventional metrics. Correctness-only GRPO rewards the final answer, whereas CoRA also penalizes cases in which confidence exceeds rationale support. As a result, CoRA may prefer responses that are better supported or less overconfident, even when a correctness-only objective would favor more aggressive predictions. This behavior is desirable when the goal is to reduce unsupported high-confidence reasoning, but it also means that confidence--rationale alignment should be viewed as complementary to answer accuracy, not as a guaranteed way to maximize it.

The domain differences suggest that the value of rationale-support supervision depends on how reliably rationale quality can be assessed. MathQA provides relatively explicit computational and logical structure, which may make rationale-support judgments more stable. MedQA requires specialized domain knowledge and may introduce greater ambiguity for both the policy model and the judge. This helps explain why the gains are broadest on MathQA and more variable on MedQA.

Evaluations under Llama 3.1 70B Instruct, GPT-OSS 20B, and Qwen 3 32B further suggest that the observed improvements are not solely an artifact of the training judge. %GPT-OSS 20B is used for reward computation during training, while GPT-OSS 20B and Llama 3 70B are applied separately during evaluation. 
The fact that CoRA often reduces the alignment error under all judges provides evidence that the method improves a broader rationale-support signal rather than only matching the preferences of a single judge. Meanwhile, disagreements between judges indicate that rationale-support evaluation remains imperfect and should not be treated as a substitute for human assessment.

Overall, our findings support treating answer correctness, confidence, and rationale support as coupled reliability signals. A reasoning model should not merely be accurate or calibrated on average; it should assign high confidence only when its rationale adequately supports its answer.

\section*{Limitations}

There are several limitations to this work that we believe serve as opportunities for further investigation and methodological advancement. First, our rationale-support score depends on LLM judges, whose judgments may contain biases or inconsistencies despite the use of a structured rubric and evaluation across multiple judge models. Future work should compare judge scores with expert and non-expert human assessments of rationale support. Second, our current experiments use relatively small training subsets, so stronger trends may require larger-scale training and multiple random seeds. Third, although the alignment error \(E_{\mathrm{align}}\) captures unsupported overconfidence, it does not fully measure all forms of rationale faithfulness. A rationale may appear supportive according to the rubric while still not reflecting the model's internal decision process. Finally, our current evaluation focuses on multiple-choice reasoning, and future work should extend confidence--rationale alignment to open-ended and retrieval-intensive reasoning tasks.
%\section*{Acknowledgments}
% This work was partially supported by a grant from the Intuit University Collaboration Program. 

\bibliography{custom}

@inproceedings{wei2022chain,
  title     = {Chain-of-Thought Prompting Elicits Reasoning in Large Language Models},
  author    = {Wei, Jason and Wang, Xuezhi and Schuurmans, Dale and Bosma, Maarten and Ichter, Brian and Xia, Fei and Chi, Ed and Le, Quoc and Zhou, Denny},
  booktitle = {Advances in Neural Information Processing Systems},
  volume    = {35},
  pages     = {24824--24837},
  year      = {2022},
  url       = {https://proceedings.neurips.cc/paper_files/paper/2022/hash/9d5609613524ecf4f15af0f7b31abca4-Abstract-Conference.html}
}

@inproceedings{kojima2022large,
  title     = {Large Language Models are Zero-Shot Reasoners},
  author    = {Kojima, Takeshi and Gu, Shixiang Shane and Reid, Machel and Matsuo, Yutaka and Iwasawa, Yusuke},
  booktitle = {Advances in Neural Information Processing Systems},
  volume    = {35},
  pages     = {22199--22213},
  year      = {2022},
  url       = {https://proceedings.neurips.cc/paper_files/paper/2022/hash/8bb0d291acd4acf06ef112099c16f326-Abstract-Conference.html}
}

@inproceedings{wang2023selfconsistency,
  title     = {Self-Consistency Improves Chain of Thought Reasoning in Language Models},
  author    = {Wang, Xuezhi and Wei, Jason and Schuurmans, Dale and Le, Quoc and Chi, Ed and Narang, Sharan and Chowdhery, Aakanksha and Zhou, Denny},
  booktitle = {International Conference on Learning Representations},
  year      = {2023},
  url       = {https://openreview.net/forum?id=1PL1NIMMrw}
}

@inproceedings{yao2023tree,
  title     = {Tree of Thoughts: Deliberate Problem Solving with Large Language Models},
  author    = {Yao, Shunyu and Yu, Dian and Zhao, Jeffrey and Shafran, Izhak and Griffiths, Thomas L. and Cao, Yuan and Narasimhan, Karthik},
  booktitle = {Advances in Neural Information Processing Systems},
  volume    = {36},
  year      = {2023},
  url       = {https://proceedings.neurips.cc/paper_files/paper/2023/hash/271db9922b8d1f4dd7aaef84ed5ac703-Abstract-Conference.html}
}

@inproceedings{turpin2023unfaithful,
  title     = {Language Models Don't Always Say What They Think: Unfaithful Explanations in Chain-of-Thought Prompting},
  author    = {Turpin, Miles and Michael, Julian and Perez, Ethan and Bowman, Samuel R.},
  booktitle = {Advances in Neural Information Processing Systems},
  volume    = {36},
  year      = {2023},
  url       = {https://proceedings.neurips.cc/paper_files/paper/2023/hash/ed3fea9033a80fea1376299fa7863f4a-Abstract-Conference.html}
}

@article{lanham2023faithfulness,
  title   = {Measuring Faithfulness in Chain-of-Thought Reasoning},
  author  = {Lanham, Tamera and Chen, Anna and Radhakrishnan, Ansh and Steiner, Benoit and Denison, Carson and Hernandez, Danny and Li, Dustin and Durmus, Esin and Hubinger, Evan and Kernion, Jackson and Luko{\v{s}}i{\=u}t{\.e}, Kamil{\.e} and Nguyen, Karina and Cheng, Newton and Joseph, Nicholas and Schiefer, Nicholas and Rausch, Oliver and Larson, Robin and McCandlish, Sam and Kundu, Sandipan and Kadavath, Saurav and Yang, Shannon and Henighan, Thomas and Maxwell, Timothy and Telleen-Lawton, Timothy and Hume, Tristan and Hatfield-Dodds, Zac and Kaplan, Jared and Brauner, Jan and Bowman, Samuel R. and Perez, Ethan},
  journal = {arXiv preprint arXiv:2307.13702},
  year    = {2023},
  url     = {https://arxiv.org/abs/2307.13702}
}

@inproceedings{paul2024making,
  title     = {Making Reasoning Matter: Measuring and Improving Faithfulness of Chain-of-Thought Reasoning},
  author    = {Paul, Debjit and West, Robert and Bosselut, Antoine and Faltings, Boi},
  booktitle = {Findings of the Association for Computational Linguistics: EMNLP 2024},
  pages     = {15012--15032},
  year      = {2024},
  address   = {Miami, Florida, USA},
  publisher = {Association for Computational Linguistics},
  doi       = {10.18653/v1/2024.findings-emnlp.882},
  url       = {https://aclanthology.org/2024.findings-emnlp.882/}
}

@inproceedings{tutek2025measuring,
  title     = {Measuring Chain of Thought Faithfulness by Unlearning Reasoning Steps},
  author    = {Tutek, Martin and Hashemi Chaleshtori, Fateme and Marasovi{\'c}, Ana and Belinkov, Yonatan},
  booktitle = {Proceedings of the 2025 Conference on Empirical Methods in Natural Language Processing},
  pages     = {9935--9960},
  year      = {2025},
  address   = {Suzhou, China},
  publisher = {Association for Computational Linguistics},
  doi       = {10.18653/v1/2025.emnlp-main.504},
  url       = {https://aclanthology.org/2025.emnlp-main.504/}
}

@inproceedings{guo2017calibration,
  title     = {On Calibration of Modern Neural Networks},
  author    = {Guo, Chuan and Pleiss, Geoff and Sun, Yu and Weinberger, Kilian Q.},
  booktitle = {Proceedings of the 34th International Conference on Machine Learning},
  series    = {Proceedings of Machine Learning Research},
  volume    = {70},
  pages     = {1321--1330},
  year      = {2017},
  publisher = {PMLR},
  url       = {https://proceedings.mlr.press/v70/guo17a.html}
}

@article{brier1950verification,
  title   = {Verification of Forecasts Expressed in Terms of Probability},
  author  = {Brier, Glenn W.},
  journal = {Monthly Weather Review},
  volume  = {78},
  number  = {1},
  pages   = {1--3},
  year    = {1950},
  doi     = {10.1175/1520-0493(1950)078<0001:VOFEIT>2.0.CO;2},
  url     = {https://journals.ametsoc.org/view/journals/mwre/78/1/1520-0493_1950_078_0001_vofeit_2_0_co_2.xml}
}

@inproceedings{zhao2021calibrate,
  title     = {Calibrate Before Use: Improving Few-Shot Performance of Language Models},
  author    = {Zhao, Tony Z. and Wallace, Eric and Feng, Shi and Klein, Dan and Singh, Sameer},
  booktitle = {Proceedings of the 38th International Conference on Machine Learning},
  series    = {Proceedings of Machine Learning Research},
  volume    = {139},
  pages     = {12697--12706},
  year      = {2021},
  publisher = {PMLR},
  url       = {https://proceedings.mlr.press/v139/zhao21c.html}
}

@article{kadavath2022language,
  title   = {Language Models (Mostly) Know What They Know},
  author  = {Kadavath, Saurav and Conerly, Tom and Askell, Amanda and Henighan, Tom and Drain, Dawn and Perez, Ethan and Schiefer, Nicholas and Hatfield-Dodds, Zac and DasSarma, Nova and Tran-Johnson, Eli and Johnston, Scott and El-Showk, Sheer and Jones, Andy and Elhage, Nelson and Hume, Tristan and Chen, Anna and Bai, Yuntao and Bowman, Sam and Fort, Stanislav and Ganguli, Deep and Hernandez, Danny and Jacobson, Josh and Kernion, Jackson and Kravec, Shauna and Lovitt, Liane and Ndousse, Kamal and Olsson, Catherine and Ringer, Sam and Amodei, Dario and Brown, Tom and Clark, Jack and Joseph, Nicholas and Mann, Ben and McCandlish, Sam and Olah, Chris and Kaplan, Jared},
  journal = {arXiv preprint arXiv:2207.05221},
  year    = {2022},
  url     = {https://arxiv.org/abs/2207.05221}
}

@inproceedings{tian2023just,
  title     = {Just Ask for Calibration: Strategies for Eliciting Calibrated Confidence Scores from Language Models Fine-Tuned with Human Feedback},
  author    = {Tian, Katherine and Mitchell, Eric and Zhou, Allan and Sharma, Archit and Rafailov, Rafael and Yao, Huaxiu and Finn, Chelsea and Manning, Christopher},
  booktitle = {Proceedings of the 2023 Conference on Empirical Methods in Natural Language Processing},
  pages     = {5433--5442},
  year      = {2023},
  address   = {Singapore},
  publisher = {Association for Computational Linguistics},
  doi       = {10.18653/v1/2023.emnlp-main.330},
  url       = {https://aclanthology.org/2023.emnlp-main.330/}
}

@inproceedings{xie2024calibrating,
  title     = {Calibrating Language Models with Adaptive Temperature Scaling},
  author    = {Xie, Johnathan and Chen, Annie S. and Lee, Yoonho and Mitchell, Eric and Finn, Chelsea},
  booktitle = {Proceedings of the 2024 Conference on Empirical Methods in Natural Language Processing},
  pages     = {18128--18138},
  year      = {2024},
  address   = {Miami, Florida, USA},
  publisher = {Association for Computational Linguistics},
  doi       = {10.18653/v1/2024.emnlp-main.1007},
  url       = {https://aclanthology.org/2024.emnlp-main.1007/}
}

@inproceedings{thermometer2024,
  title     = {Thermometer: Towards Universal Calibration for Large Language Models},
  author    = {Shen, Maohao and Das, Subhro and Greenewald, Kristjan and Sattigeri, Prasanna and Wornell, Gregory and Ghosh, Soumya},
  booktitle = {Proceedings of the 41st International Conference on Machine Learning},
  series    = {Proceedings of Machine Learning Research},
  volume    = {235},
  pages     = {44687--44711},
  year      = {2024},
  publisher = {PMLR},
  url       = {https://proceedings.mlr.press/v235/shen24c.html}
}

@inproceedings{geng2024survey,
  title     = {A Survey of Confidence Estimation and Calibration in Large Language Models},
  author    = {Geng, Jiahui and Cai, Fengyu and Wang, Yuxia and Koeppl, Heinz and Nakov, Preslav and Gurevych, Iryna},
  booktitle = {Proceedings of the 2024 Conference of the North American Chapter of the Association for Computational Linguistics: Human Language Technologies (Volume 1: Long Papers)},
  pages     = {6577--6595},
  year      = {2024},
  address   = {Mexico City, Mexico},
  publisher = {Association for Computational Linguistics},
  doi       = {10.18653/v1/2024.naacl-long.366},
  url       = {https://aclanthology.org/2024.naacl-long.366/}
}

@inproceedings{zheng2023judging,
  title     = {Judging {LLM}-as-a-Judge with {MT}-Bench and Chatbot Arena},
  author    = {Zheng, Lianmin and Chiang, Wei-Lin and Sheng, Ying and Zhuang, Siyuan and Wu, Zhanghao and Zhuang, Yonghao and Lin, Zi and Li, Zhuohan and Li, Dacheng and Xing, Eric P. and Zhang, Hao and Gonzalez, Joseph E. and Stoica, Ion},
  booktitle = {Advances in Neural Information Processing Systems},
  volume    = {36},
  year      = {2023},
  url       = {https://proceedings.neurips.cc/paper_files/paper/2023/hash/91f18a1287b398d378ef22505bf41832-Abstract-Datasets_and_Benchmarks.html}
}

@inproceedings{liu2023geval,
  title     = {G-Eval: {NLG} Evaluation using {GPT}-4 with Better Human Alignment},
  author    = {Liu, Yang and Iter, Dan and Xu, Yichong and Wang, Shuohang and Xu, Ruochen and Zhu, Chenguang},
  booktitle = {Proceedings of the 2023 Conference on Empirical Methods in Natural Language Processing},
  pages     = {2511--2522},
  year      = {2023},
  address   = {Singapore},
  publisher = {Association for Computational Linguistics},
  doi       = {10.18653/v1/2023.emnlp-main.153},
  url       = {https://aclanthology.org/2023.emnlp-main.153/}
}

@inproceedings{kim2024prometheus,
  title     = {Prometheus: Inducing Fine-Grained Evaluation Capability in Language Models},
  author    = {Kim, Seungone and Shin, Jamin and Cho, Yejin and Jang, Joel and Longpre, Shayne and Lee, Hwaran and Yun, Sangdoo and Shin, Seongjin and Kim, Sungdong and Thorne, James and Seo, Minjoon},
  booktitle = {International Conference on Learning Representations},
  year      = {2024},
  url       = {https://openreview.net/forum?id=8euJaTveKw}
}

@inproceedings{conqord2024,
  title     = {When to Trust {LLM}s: Aligning Confidence with Response Quality},
  author    = {Tao, Shuchang and Yao, Liuyi and Ding, Hanxing and Xie, Yuexiang and Cao, Qi and Sun, Fei and Gao, Jinyang and Shen, Huawei and Ding, Bolin},
  booktitle = {Findings of the Association for Computational Linguistics: ACL 2024},
  pages     = {5984--5996},
  year      = {2024},
  address   = {Bangkok, Thailand},
  publisher = {Association for Computational Linguistics},
  doi       = {10.18653/v1/2024.findings-acl.357},
  url       = {https://aclanthology.org/2024.findings-acl.357/}
}

@inproceedings{zhang2025cot,
  title     = {{CoT}-{UQ}: Improving Response-wise Uncertainty Quantification in {LLM}s with Chain-of-Thought},
  author    = {Zhang, Boxuan and Zhang, Ruqi},
  booktitle = {Findings of the Association for Computational Linguistics: ACL 2025},
  pages     = {26114--26133},
  month     = jul,
  year      = {2025},
  address   = {Vienna, Austria},
  publisher = {Association for Computational Linguistics},
  doi       = {10.18653/v1/2025.findings-acl.1339},
  url       = {https://aclanthology.org/2025.findings-acl.1339/}
}

@inproceedings{razghandi2025cer,
  title     = {{CER}: Confidence Enhanced Reasoning in {LLM}s},
  author    = {Razghandi, Ali and Hosseini, Seyed Mohammad Hadi and Baghshah, Mahdieh Soleymani},
  booktitle = {Proceedings of the 63rd Annual Meeting of the Association for Computational Linguistics (Volume 1: Long Papers)},
  pages     = {7918--7938},
  month     = jul,
  year      = {2025},
  address   = {Vienna, Austria},
  publisher = {Association for Computational Linguistics},
  doi       = {10.18653/v1/2025.acl-long.390},
  url       = {https://aclanthology.org/2025.acl-long.390/}
}

@article{steyvers2025what,
  title   = {What Large Language Models Know and What People Think They Know},
  author  = {Steyvers, Mark and Tejeda, Heliodoro and Kumar, Aakriti and Belem, Catarina and Karny, Sheer and Hu, Xinyue and Mayer, Lukas and Smyth, Padhraic},
  journal = {Nature Machine Intelligence},
  volume  = {7},
  pages   = {221--231},
  year    = {2025},
  doi     = {10.1038/s42256-024-00976-7},
  url     = {https://www.nature.com/articles/s42256-024-00976-7}
}

@inproceedings{kim2025fostering,
  title     = {Fostering Appropriate Reliance on Large Language Models: The Role of Explanations, Sources, and Inconsistencies},
  author    = {Kim, Sunnie S. Y. and Vaughan, Jennifer Wortman and Liao, Q. Vera and Lombrozo, Tania and Russakovsky, Olga},
  booktitle = {Proceedings of the 2025 CHI Conference on Human Factors in Computing Systems},
  year      = {2025},
  address   = {Yokohama, Japan},
  publisher = {Association for Computing Machinery},
  numpages  = {19},
  doi       = {10.1145/3706598.3714020},
  url       = {https://doi.org/10.1145/3706598.3714020}
}

@article{vasconcelos2023explanations,
  title     = {Explanations Can Reduce Overreliance on {AI} Systems During Decision-Making},
  author    = {Vasconcelos, Helena and J{\"o}rke, Matthew and Grunde-McLaughlin, Madeleine and Gerstenberg, Tobias and Bernstein, Michael S. and Krishna, Ranjay},
  journal   = {Proceedings of the ACM on Human-Computer Interaction},
  volume    = {7},
  number    = {CSCW1},
  articleno = {129},
  pages     = {1--38},
  year      = {2023},
  publisher = {Association for Computing Machinery},
  doi       = {10.1145/3579605},
  url       = {https://doi.org/10.1145/3579605}
}

@article{shao2024deepseekmath,
  title   = {DeepSeekMath: Pushing the Limits of Mathematical Reasoning in Open Language Models},
  author  = {Shao, Zhihong and Wang, Peiyi and Zhu, Qihao and Xu, Runxin and Song, Junxiao and Bi, Xiao and Zhang, Haowei and Zhang, Mingchuan and Li, Y. K. and Wu, Y. and Guo, Daya},
  journal = {arXiv preprint arXiv:2402.03300},
  year    = {2024},
  url     = {https://arxiv.org/abs/2402.03300}
}

@article{jin2021medqa,
  title     = {What Disease Does This Patient Have? A Large-Scale Open Domain Question Answering Dataset from Medical Exams},
  author    = {Jin, Di and Pan, Eileen and Oufattole, Nassim and Weng, Wei-Hung and Fang, Hanyi and Szolovits, Peter},
  journal   = {Applied Sciences},
  volume    = {11},
  number    = {14},
  pages     = {6421},
  year      = {2021},
  publisher = {MDPI},
  doi       = {10.3390/app11146421},
  url       = {https://www.mdpi.com/2076-3417/11/14/6421}
}

@inproceedings{amini2019mathqa,
  title     = {MathQA: Towards Interpretable Math Word Problem Solving with Operation-Based Formalisms},
  author    = {Amini, Aida and Gabriel, Saadia and Lin, Shanchuan and Koncel-Kedziorski, Rik and Choi, Yejin and Hajishirzi, Hannaneh},
  booktitle = {Proceedings of the 2019 Conference of the North American Chapter of the Association for Computational Linguistics: Human Language Technologies},
  pages     = {2357--2367},
  year      = {2019},
  address   = {Minneapolis, Minnesota},
  publisher = {Association for Computational Linguistics},
  doi       = {10.18653/v1/N19-1245},
  url       = {https://aclanthology.org/N19-1245/}
}

@inproceedings{mihaylov2018openbookqa,
  title     = {Can a Suit of Armor Conduct Electricity? A New Dataset for Open Book Question Answering},
  author    = {Mihaylov, Todor and Clark, Peter and Khot, Tushar and Sabharwal, Ashish},
  booktitle = {Proceedings of the 2018 Conference on Empirical Methods in Natural Language Processing},
  pages     = {2381--2391},
  year      = {2018},
  address   = {Brussels, Belgium},
  publisher = {Association for Computational Linguistics},
  doi       = {10.18653/v1/D18-1260},
  url       = {https://aclanthology.org/D18-1260/}
}

@misc{qwen2025qwen25technicalreport,
  title         = {Qwen2.5 Technical Report},
  author        = {{Qwen Team}},
  year          = {2024},
  eprint        = {2412.15115},
  archivePrefix = {arXiv},
  primaryClass  = {cs.CL},
  url           = {https://arxiv.org/abs/2412.15115}
}

@misc{gemmateam2024gemma2,
  title         = {Gemma 2: Improving Open Language Models at a Practical Size},
  author        = {{Gemma Team}},
  year          = {2024},
  eprint        = {2408.00118},
  archivePrefix = {arXiv},
  primaryClass  = {cs.CL},
  url           = {https://arxiv.org/abs/2408.00118}
}

@inproceedings{lei2016rationalizing,
  title     = {Rationalizing Neural Predictions},
  author    = {Lei, Tao and Barzilay, Regina and Jaakkola, Tommi},
  booktitle = {Proceedings of the 2016 Conference on Empirical Methods in Natural Language Processing},
  pages     = {107--117},
  year      = {2016},
  address   = {Austin, Texas},
  publisher = {Association for Computational Linguistics},
  doi       = {10.18653/v1/D16-1011},
  url       = {https://aclanthology.org/D16-1011/}
}

@inproceedings{jacovi2020towards,
  title     = {Towards Faithfully Interpretable {NLP} Systems: How Should We Define and Evaluate Faithfulness?},
  author    = {Jacovi, Alon and Goldberg, Yoav},
  booktitle = {Proceedings of the 58th Annual Meeting of the Association for Computational Linguistics},
  pages     = {4198--4205},
  year      = {2020},
  address   = {Online},
  publisher = {Association for Computational Linguistics},
  doi       = {10.18653/v1/2020.acl-main.386},
  url       = {https://aclanthology.org/2020.acl-main.386/}
}

@inproceedings{deyoung2020eraser,
  title     = {{ERASER}: A Benchmark to Evaluate Rationalized {NLP} Models},
  author    = {DeYoung, Jay and Jain, Sarthak and Rajani, Nazneen Fatema and Lehman, Eric and Xiong, Caiming and Socher, Richard and Wallace, Byron C.},
  booktitle = {Proceedings of the 58th Annual Meeting of the Association for Computational Linguistics},
  pages     = {4443--4458},
  year      = {2020},
  address   = {Online},
  publisher = {Association for Computational Linguistics},
  doi       = {10.18653/v1/2020.acl-main.408},
  url       = {https://aclanthology.org/2020.acl-main.408/}
}

@article{lin2022teaching,
  title   = {Teaching Models to Express Their Uncertainty in Words},
  author  = {Lin, Stephanie and Hilton, Jacob and Evans, Owain},
  journal = {arXiv preprint arXiv:2205.14334},
  year    = {2022},
  url     = {https://arxiv.org/abs/2205.14334}
}

@inproceedings{zelikman2022star,
  title     = {{STaR}: Bootstrapping Reasoning with Reasoning},
  author    = {Zelikman, Eric and Wu, Yuhuai and Mu, Jesse and Goodman, Noah D.},
  booktitle = {Advances in Neural Information Processing Systems},
  volume    = {35},
  pages     = {15476--15488},
  year      = {2022},
  url       = {https://proceedings.neurips.cc/paper_files/paper/2022/hash/639a9a172c044fbb64175b5fad42e9a5-Abstract-Conference.html}
}

@article{lightman2023lets,
  title   = {Let's Verify Step by Step},
  author  = {Lightman, Hunter and Kosaraju, Vineet and Burda, Yura and Edwards, Harri and Baker, Bowen and Lee, Teddy and Leike, Jan and Schulman, John and Sutskever, Ilya and Cobbe, Karl},
  journal = {arXiv preprint arXiv:2305.20050},
  year    = {2023},
  url     = {https://arxiv.org/abs/2305.20050}
}

@misc{deepseekai2025deepseekr1arxiv,
  title         = {DeepSeek-R1: Incentivizing Reasoning Capability in {LLM}s via Reinforcement Learning},
  author        = {{DeepSeek-AI}},
  year          = {2025},
  eprint        = {2501.12948},
  archivePrefix = {arXiv},
  primaryClass  = {cs.CL},
  url           = {https://arxiv.org/abs/2501.12948}
}

@article{schulman2017proximal,
  title   = {Proximal Policy Optimization Algorithms},
  author  = {Schulman, John and Wolski, Filip and Dhariwal, Prafulla and Radford, Alec and Klimov, Oleg},
  journal = {arXiv preprint arXiv:1707.06347},
  year    = {2017},
  url     = {https://arxiv.org/abs/1707.06347}
}

@inproceedings{ouyang2022training,
  title     = {Training Language Models to Follow Instructions with Human Feedback},
  author    = {Ouyang, Long and Wu, Jeffrey and Jiang, Xu and Almeida, Diogo and Wainwright, Carroll L. and Mishkin, Pamela and Zhang, Chong and Agarwal, Sandhini and Slama, Katarina and Ray, Alex and Schulman, John and Hilton, Jacob and Kelton, Fraser and Miller, Luke and Simens, Maddie and Askell, Amanda and Welinder, Peter and Christiano, Paul and Leike, Jan and Lowe, Ryan},
  booktitle = {Advances in Neural Information Processing Systems},
  volume    = {35},
  pages     = {27730--27744},
  year      = {2022},
  url       = {https://proceedings.neurips.cc/paper_files/paper/2022/hash/b1efde53be364a73914f58805a001731-Abstract-Conference.html}
}

@misc{yang2025qwen3technicalreport,
  title         = {Qwen3 Technical Report},
  author        = {Yang, An and Li, Anfeng and Yang, Baosong and Zhang, Beichen and Hui, Binyuan and Zheng, Bo and Yu, Bowen and Gao, Chang and Huang, Chengen and Lv, Chenxu and Zheng, Chujie and Liu, Dayiheng and Zhou, Fan and Huang, Fei and Hu, Feng and Ge, Hao and Wei, Haoran and Lin, Huan and Tang, Jialong and Yang, Jian and Tu, Jianhong and Zhang, Jianwei and Yang, Jianxin and Yang, Jiaxi and Zhou, Jing and Zhou, Jingren and Lin, Junyang and Dang, Kai and Bao, Keqin and Yang, Kexin and Yu, Le and Deng, Lianghao and Li, Mei and Li, Mingze and Zhang, Pei and Wang, Peng and Zhu, Qin and Men, Rui and Gao, Ruize and Liu, Shixuan and Luo, Shuang and Li, Tianhao and Tang, Tianyi and Yin, Wenbiao and Ren, Xingzhang and Wang, Xinyu and Zhang, Xinyu and Ren, Xuancheng and Fan, Yang and Su, Yang and Zhang, Yichang and Zhang, Yinger and Wan, Yu and Liu, Yuqiong and Wang, Zekun and Cui, Zeyu and Zhang, Zhenru and Zhou, Zhipeng and Qiu, Zihan},
  year          = {2025},
  eprint        = {2505.09388},
  archivePrefix = {arXiv},
  primaryClass  = {cs.CL},
  url           = {https://arxiv.org/abs/2505.09388}
}

@misc{mistral2025ministral3reasoning,
  title        = {Ministral 3 8B Reasoning 2512},
  author       = {{Mistral AI}},
  year         = {2025},
  howpublished = {Hugging Face Model Card},
  url          = {https://huggingface.co/mistralai/Ministral-3-8B-Reasoning-2512}
}

@article{agarwal2025gpt,
  title = {gpt-oss-120b \& gpt-oss-20b Model Card},
  author = {{OpenAI}},
  journal = {arXiv preprint arXiv:2508.10925},
  year = {2025},
  url = {https://arxiv.org/abs/2508.10925}
}

@article{dubey2024llama,
  title = {The Llama 3 Herd of Models},
  author = {{Llama Team, AI @ Meta}},
  journal = {arXiv preprint arXiv:2407.21783},
  year = {2024},
  url = {https://arxiv.org/abs/2407.21783}
}

\appendix

\section{Additional Implementation Details}
\label{app:implementation}

\subsection{Committed-Answer Confidence Extraction}
\label{app:confidence-extraction}

For each generated completion, we first parse the committed answer \(y\) from the final answer line \texttt{Final Answer: X}. We then compute confidence over the discrete answer-option labels rather than using the model's verbalized confidence. Specifically, after the model generates a rationale and a final-answer line, we locate the generated answer label in the token sequence and take the exact generated prefix before that label as the scoring context. Under this prefix, we score each candidate answer label \(\ell \in \mathcal{Y}\) as a continuation.

In our implementation, candidate labels are scored with a leading-space template \texttt{\space X}, where \(X\) is one of the valid option labels. Thus, for a four-choice dataset, the candidate continuations are \texttt{\space A}, \texttt{\space B}, \texttt{\space C}, and \texttt{\space D}. For MedQA and MathQA, we additionally include \texttt{\space E}. Let \(s_\ell\) denote the conditional log-probability of candidate continuation \(\ell\) under the exact generated prefix. We convert these scores into an option-level probability distribution:
\[
p(\ell \mid x,r) =
\frac{\exp(s_\ell)}
{\sum_{\ell' \in \mathcal{Y}}\exp(s_{\ell'})}.
\]
The committed-answer confidence is then defined as
\[
C = p(y \mid x,r),
\]
where \(y\) is the parsed answer selected by the model. Since the candidate continuations are fixed answer labels rather than full option texts, they are short and comparable across options; we therefore use the summed continuation log-probability without additional length normalization.

No temperature scaling is applied when computing these option-level probabilities. Although generation may use sampling parameters during response generation, the confidence score is computed post hoc from the model's raw conditional likelihoods under the generated prefix. The resulting distribution is stored as \texttt{choice\_probs}, and the probability assigned to the parsed answer is stored as \texttt{p\_parsed}. This \texttt{p\_parsed} value is used as \(C\in[0,1]\) for ECE, Brier score, the confidence--rationale alignment reward, and the alignment error metric. If a completion does not contain a parseable final answer, it is counted as incorrect and assigned zero rationale-support score. This procedure measures the model's probability for the committed answer label under the generated reasoning context, rather than the probability of the full option text or any self-reported confidence statement.

\subsection{Reproducibility Details}
\label{app:reproducibility}
For reward computation during training, we use GPT-OSS 20B as the sole rationale-support judge. The judge input excludes gold option labels, official solutions, official explanations, and gold rationales; the auxiliary background knowledge is used only to assess factual grounding. The general support score \(Q_{\mathrm{gen}}\) is computed from the rubric axes in Table~\ref{tab:rubric}. The format-validity axis \(G_1\) is used as a hard gate. For the remaining general axes, we use weights of 0.20, 0.30, 0.20, 0.25, and 0.05 for task understanding, evidence grounding, inference coherence, answer bridge, and structure, respectively. The task-specific score \(Q_{\mathrm{task}}\) is computed by averaging the two task-specific axes. We set \(w_{\mathrm{gen}}=0.75\) and \(w_{\mathrm{task}}=0.25\), apply the penalty \(P\) for severe structural failure or unsupported reasoning, and clip the final score to \([0,1]\). The training process uses 3 NVIDIA H100 GPUs.

For the confidence--rationale alignment score in Equation~\ref{eq:t-score}, we set \(\alpha=0.6\), \(\beta_{\mathrm{over}}=0.6\), and \(\beta_{\mathrm{under}}=0.4\), assigning a larger penalty to confidence that exceeds rationale support. For the final reward in Equation~\ref{eq:reward}, we use \(\lambda=0.2\), \(\gamma=0.2\), and \(\beta=0.1\). We train with LoRA using rank 16, LoRA alpha 32, and dropout 0.05. Each model is trained for one epoch with learning rate \(10^{-5}\), batch size 8, group size 8 for GRPO sampling, and KL coefficient 0.04.

\subsection{Prompt Templates}
\label{app:prompt}

\paragraph{Generation Prompt.}
The policy model is prompted with the following template:
\begin{quote}
\scriptsize
\begin{verbatim}
You are a careful reasoner.

Answer the following {dataset_display}
multiple-choice question. Give a short
chain-of-thought in several steps.
Then output Final Answer as a single
letter from {choice_list}. Stop immediately
after the Final Answer line.

{fact_block}Question: {question}
Options:
{options_text}

Use exactly this format:
Chain-of-Thought:
1) ...
2) ...
Final Answer: {choice_template}

\end{verbatim}
\end{quote}

% Here, \texttt{\{dataset\_display\}} is instantiated as the dataset-specific task description, \texttt{\{choice\_list\}} denotes the available answer labels, and \texttt{\{choice\_template\}} denotes the valid answer-label format. The optional \texttt{\{fact\_block\}} is included only when reference evidence is provided to the generation model.

\paragraph{Judge System Prompt.}
The rubric judge is prompted with the following system instruction:
\begin{quote}
\scriptsize
\begin{verbatim}
Reasoning: low

You are an evaluator for multiple-choice
question reasoning. You are given:
(a) a dataset name
(b) a question and answer options
(c) optional reference/evidence
(d) a CHAIN-OF-THOUGHT answer written
    by another model

Score the CHAIN along the 8 rubric axes.

Use reliable domain knowledge when needed:
- OpenBookQA: science/common-sense knowledge
- MedQA: medical knowledge
- MathQA: mathematical reasoning/computation

You are NOT given the correct answer.
Do not use a gold label. Judge only from
the question, options, optional evidence,
and chain.

For each axis, output the value only.
No evidence quotes. Output STRICT JSON only.
\end{verbatim}
\end{quote}

\paragraph{Judge Input Prompt.}
For each generated reasoning trace, the judge receives:
\begin{quote}
\scriptsize
\begin{verbatim}
DATASET: {dataset_name}

QUESTION: {question}

OPTIONS:
A: {option_A}
B: {option_B}
...

PROVIDED REFERENCE/EVIDENCE:
{fact1}

CHAIN-OF-THOUGHT:
{chain}

CHOSEN ANSWER: {parsed_answer}

Evaluate this chain along the 8 rubric axes.
Output STRICT JSON only.
\end{verbatim}
\end{quote}

The reference-evidence block is omitted when no reference evidence is used.

\section{Linguistic Features for Correctness Prediction}
\label{app:linguistic-feature}

\begin{table}[t]
\centering
\caption{Linguistic features used for correctness prediction. Features are extracted from the generated CoT reasoning trace.}
\scriptsize
\setlength{\tabcolsep}{4pt}
\renewcommand{\arraystretch}{1.00}
\resizebox{\columnwidth}{!}{
\begin{tabular}{ll}
\toprule
\textbf{Feature} & \textbf{Description} \\
\midrule
Length & Number of characters and words in the generated trace. \\
Step count & Number of explicit numbered reasoning steps. \\
Sentence length & Average number of words per sentence. \\
Hedging & Density of uncertainty markers, e.g., ``maybe'', ``likely''. \\
Connectives & Density of causal/logical connectives, e.g., ``therefore''. \\
Backtracking & Density of revision markers, e.g., ``wait'', ``actually''. \\
Digit density & Fraction of characters that are digits. \\
Repetition & Duplicate trigram rate in the generated trace. \\
Lexical diversity & Ratio of unique words to total words. \\
\bottomrule
\end{tabular}
}
\label{tab:linguistic-features}
\end{table}
% \FloatBarrier

\section{Additional Judge Evaluation}
\label{app:judge-robustness}

\begin{table*}[t]
\centering

\caption{\textbf{Rationale-support quality and confidence--rationale alignment results evaluated by GPT-OSS 20B}. \(Q\) measures the quality of the generated reasoning trace. \(E_{\mathrm{align}}\) denotes the average confidence--rationale alignment error. \(\uparrow\) indicates higher is better, and \(\downarrow\) indicates lower is better. All values are reported as percentages.}
\scriptsize
\setlength{\tabcolsep}{4pt}
\renewcommand{\arraystretch}{1.0}
\resizebox{\textwidth}{!}{
\begin{tabular}{llcccccc}
\toprule
\multicolumn{2}{c}{} 
& \multicolumn{2}{c}{\textbf{Ministral 3 8B}}
& \multicolumn{2}{c}{\textbf{Qwen2.5 7B}}
& \multicolumn{2}{c}{\textbf{Gemma 2 9B}} \\
\cmidrule(lr){3-4} \cmidrule(lr){5-6} \cmidrule(lr){7-8}
\textbf{Dataset} & \textbf{Method}
& \(Q\) $\uparrow$ & \(E_{\mathrm{align}}\) $\downarrow$
& \(Q\) $\uparrow$ & \(E_{\mathrm{align}}\) $\downarrow$
& \(Q\) $\uparrow$ & \(E_{\mathrm{align}}\) $\downarrow$ \\
\midrule
\multirow{4}{*}{MathQA}
& Base & 94.51 & 6.72 & 83.67 & 13.61 & 75.01 & 21.11 \\
& SFT  & 94.77 & 5.42 & 82.41 & 14.29 & \second{76.75} & 21.20 \\
& GRPO & \second{96.19} & \second{5.14} & \second{84.93} & \second{11.50} & 76.59 & \second{20.06} \\
& CoRA (Ours) & \best{97.09} & \best{2.97} & \best{87.70} & \best{9.54} & \best{79.07} & \best{16.75} \\
\midrule
\multirow{4}{*}{MedQA}
& Base & 81.31 & 18.15 & 59.30 & 36.01 & 69.83 & 29.28 \\
& SFT  & 81.92 & 17.48 & 61.26 & 34.79 & 71.54 & 28.48 \\
& GRPO & \best{83.89} & \best{16.28} & \best{64.89} & \best{28.56} & \second{72.01} & \second{26.84} \\
& CoRA (Ours) & \second{82.17} & \second{16.87} & \second{63.78} & \second{32.29} & \best{78.72} & \best{20.42} \\
\midrule
\multirow{4}{*}{OpenBookQA}
& Base & 82.17 & 5.98 & 90.89 & 9.97 & 92.02 & 7.25 \\
& SFT  & \second{95.08} & \second{4.97} & \second{91.38} & \second{8.43} & 91.90 & 7.77 \\
& GRPO & 94.81 & 5.51 & 90.95 & 9.43 & \second{93.10} & \second{6.59} \\
& CoRA (Ours) & \best{95.48} & \best{4.24} & \best{92.41} & \best{7.73} & \best{93.25} & \best{6.15} \\
\bottomrule
\end{tabular}
}

\label{tab:q-align-results-gpt}
\end{table*}

\begin{table*}[t]
\centering
\caption{\textbf{Rationale-support quality and confidence--rationale alignment results evaluated by Qwen 3 32B.} \(Q\) measures the quality of the generated reasoning trace, while \(E_{\mathrm{align}}\) denotes the average confidence--rationale alignment error. \(\uparrow\) indicates higher is better, and \(\downarrow\) indicates lower is better. All values are reported as percentages.}
\scriptsize
\setlength{\tabcolsep}{4pt}
\renewcommand{\arraystretch}{1.05}
\resizebox{\textwidth}{!}{
\begin{tabular}{llcccccc}
\toprule
\multicolumn{2}{c}{} 
& \multicolumn{2}{c}{\textbf{Ministral 3 8B}}
& \multicolumn{2}{c}{\textbf{Qwen2.5 7B}}
& \multicolumn{2}{c}{\textbf{Gemma 2 9B}} \\
\cmidrule(lr){3-4} \cmidrule(lr){5-6} \cmidrule(lr){7-8}
\textbf{Dataset} & \textbf{Method}
& \(Q\) $\uparrow$ & \(E_{\mathrm{align}}\) $\downarrow$
& \(Q\) $\uparrow$ & \(E_{\mathrm{align}}\) $\downarrow$
& \(Q\) $\uparrow$ & \(E_{\mathrm{align}}\) $\downarrow$ \\
\midrule
\multirow{4}{*}{MathQA}
& Base & \second{96.51} & 5.10 & 84.99 & 12.49 & 78.83 & 18.54 \\
& SFT  & 95.50 & \second{4.62} & 84.50 & 12.77 & \second{80.24} & \second{17.97} \\
& GRPO & 96.42 & 5.32 & \second{86.59} & \second{10.43} & 79.26 & 18.10 \\
& CoRA (Ours) & \best{98.68} & \best{1.31} & \best{88.33} & \best{9.49} & \best{82.98} & \best{14.29} \\
\midrule
\multirow{4}{*}{MedQA}
& Base & 88.84 & 11.41 & 67.23 & 30.98 & 72.12 & 27.29 \\
& SFT  & 89.51 & 10.39 & 70.57 & 27.83 & 74.08 & 25.86 \\
& GRPO & \best{90.99} & \best{9.35} & \second{71.12} & \best{24.08} & \second{74.63} & \second{24.57} \\
& CoRA (Ours) & \second{90.46} & \second{10.13} & \best{72.36} & \second{25.97} & \best{84.09} & \best{15.70} \\
\midrule
\multirow{4}{*}{OpenBookQA}
& Base & 93.37 & 7.44 & 91.12 & 10.14 & 89.68 & 9.94 \\
& SFT  & \best{95.22} & \best{4.72} & \second{92.24} & \second{8.22} & 90.73 & 9.36 \\
& GRPO & \second{95.04} & 5.59 & 91.36 & 9.16 & \second{91.76} & \second{8.14} \\
& CoRA (Ours) & 94.38 & \second{5.30} & \best{92.25} & \best{7.68} & \best{92.96} & \best{7.08} \\
\bottomrule
\end{tabular}
}
\label{tab:q-align-results-qwen3}
\end{table*}

\end{document}